\documentclass[journal]{IEEEtran}
\usepackage{amsmath}
\usepackage{indentfirst}
\usepackage{booktabs}
\usepackage{subfigure}
\usepackage{epsfig}
\usepackage{amssymb}
\usepackage{graphicx}
\usepackage{float}
\usepackage{setspace}
\usepackage{tabularx,booktabs}
\usepackage{url}
\usepackage{amsfonts,amssymb}
\usepackage{dsfont}
\usepackage{enumerate}
\usepackage{algorithmic}
\usepackage[linesnumbered,ruled]{algorithm2e}
\usepackage{stfloats}
\usepackage[numbers,sort&compress]{natbib}
\usepackage{multirow}
\usepackage{rotating}
\usepackage{sverb, longtable}
\ifCLASSINFOpdf
\else
\fi
\begin{document}
%
\title{Evidential distance measure in complex belief function theory}
%
%
%
%

\author{Fuyuan Xiao
\IEEEcompsocitemizethanks{\IEEEcompsocthanksitem F. Xiao is with the School of Computer and Information Science, Southwest University, No.2 Tiansheng Road, BeiBei District, Chongqing, 400715, China.\protect\\
E-mail: xiaofuyuan@swu.edu.cn
}
}

%
%

\markboth{}%
{
}
%



\IEEEtitleabstractindextext{%
\begin{abstract}
In this paper, an evidential distance measure is proposed which can measure the difference or dissimilarity between complex basic belief assignments (CBBAs), in which the CBBAs are composed of complex numbers.
When the CBBAs are degenerated from complex numbers to real numbers, i.e., BBAs, the proposed distance will degrade into the Jousselme et al.'s distance.
Therefore, the proposed distance provides a promising way to measure the differences between evidences in a more general framework of complex plane space.
\end{abstract}

}

\maketitle

\IEEEdisplaynontitleabstractindextext

%
\IEEEpeerreviewmaketitle

\section{Introduction}\label{Introduction}
A recent work in terms of the generalization of Dempster--Shafer evidence (GDSE) theory is presented where a new concept of complex belief function is defined based on the complex numbers~\cite{Xiao2018Generalization}.
The GDSE theory is capable of giving expression to the data fluctuations at a given time phase in the course of execution.
Moreover, it has the ability to handle uncertainty and imprecision when the data occur concurrently accompanied by variations against to data's phase or periodicity.
In particular, when the complex basic belief assignments are degenerated from complex numbers to real numbers, the GDSE theory will degrade into the DSE theory with the condition that the conflict coefficient is less than one.
Therefore, the GDSE theory can provide a more promising framework to model and cope with uncertain information.

Because the evidential distance plays an important role to measure the difference or dissimilarity between evidences in DSE theory which has attracted many researchers in the past few years.
In this paper, therefore, inspired by Jousselme's distance~\cite{jousselme2001new}, an evidential distance measure, called EDM is proposed that can measure the differences between complex basic belief assignments (CBBAs) in the GDSE theory, in which the CBBAs are composed of complex numbers.
When the CBBAs are degenerated from complex numbers to real numbers, the EDM distance will degrade into the Jousselme et al.'s distance.
Hence, the proposed EDM distance is a generalization of the Jousselme et al.'s distance.
Meanwhile, the properties of the EDM distance measure are analysed.
Furthermore, numerical examples are given to illustrate the properties of the EDM distance measure.

The rest of this paper is organised as follows.
The preliminaries are briefly introduced in Section~\ref{Preliminaries}.
A complex basic belief assignment is introduced in Section~\ref{CBBA}.
A new distance measure between complex basic belief assignments is proposed in Section~\ref{Proposed method_EDM}.
Section~\ref{Numericalexamples} provides many numerical examples to illustrate the properties of the EDM distance measure.
Finally, Section~\ref{Conclusion} concludes this work.

\section{Preliminaries}\label{Preliminaries}

\newtheorem{myDef}{Definition}
\newtheorem{exmp}{Example}
\newtheorem{Property}{Property}
\newtheorem{Remark}{Remark}
\newtheorem{Proof}{Proof}
\newtheorem{Theorem}{Theorem}

\subsection{Complex number~\cite{ablowitz2003complex}}\label{Complexnumber}
A complex number $z$ is defined as an ordered pair of real numbers
\begin{equation}\label{eq_complexnumber}
z = x + yi,
\end{equation}
where $x$ and $y$ are real numbers and $i$ is the imaginary unit, satisfying $i^2 = -1$.
This is called the ``rectangular'' form or ``Cartesian'' form.

It can also expressed in polar form, denoted by
\begin{equation}\label{eq_complexnumber}
z=r e^{i \theta},
\end{equation}
where $r > 0$ represents the modulus or magnitude of the complex number $z$ and $\theta$ represents the angle or phase of the complex number $z$.

By using the Euler's relation,
\begin{equation}\label{eq_Euler'srelation}
e^{i \theta} = \cos(\theta) +i \sin(\theta),
\end{equation}
the modulus or magnitude and angle or phase of the complex number can be expressed as
\begin{equation}\label{eq_magnitudeandphase}
r=\sqrt{x^2+y^2}, \text{ and } \theta = \arctan(\frac{y}{x}) = \tan^{-1}(\frac{y}{x}),
\end{equation}
where $x = r \cos(\theta)$ and $y = r \sin(\theta)$.

The square of the absolute value is defined by
\begin{equation}\label{eq_squaremagnitude}
|z|^2=z\bar{z}=x^2+y^2,
\end{equation}
where $\bar{z}$ is the complex conjugate of $z$, i.e., $\bar{z}=x - yi$.

These relationships can be then obtained as
\begin{equation}\label{eq_relationship}
r=|z|, \text{ and } \theta = \angle z,
\end{equation}
where if $z$ is a real number (i.e., $y = 0$), then $r = |x|$.

The arithmetic of complex numbers is defined as follows.

Give two complex numbers $z_1=x_1 + y_1i$ and $z_2=x_2 + y_2i$,

\begin{itemize}
\item
The addition is defined by
\begin{equation}\label{eq_addition}
z_1+z_2=(x_1 + y_1i)+(x_2 + y_2i)=(x_1+x_2)+(y_1+y_2)i.
\end{equation}

\item
The subtraction is defined by
\begin{equation}\label{eq_subtraction}
z_1-z_2=(x_1 + y_1i)-(x_2 + y_2i)=(x_1-x_2)+(y_1-y_2)i.
\end{equation}

\item
The multiplication is defined by
\begin{equation}\label{eq_multiplication}
(x_1 + y_1i)(x_2 + y_2i)=(x_1x_2-y_1y_2)+(x_1y_2+x_2y_1)i.
\end{equation}

\end{itemize}

\subsection{Belief function theory~\cite{Dempster1967Upper,shafer1976mathematical}}\label{BBA}

\begin{myDef}(Frame of discernment)

Let $\Omega$ be a set of mutually exclusive and collective non-empty events, defined by
\begin{equation}\label{eq_Frameofdiscernment1}
 \Omega = \{F_{1}, F_{2}, \ldots, F_{i}, \ldots, F_{N}\},
\end{equation}
where $\Omega$ is a frame of discernment.

The power set of $\Omega$ is denoted as $2^{\Omega}$,
\begin{equation}\label{eq_Frameofdiscernment2}
\begin{aligned}
 2^{\Omega} = \{\emptyset, \{F_{1}\}, \{F_{2}\}, \ldots, \{F_{N}\}, \{F_{1}, F_{2}\}, \ldots, \{F_{1}, \\
 F_{2}, \ldots, F_{i}\}, \ldots, \Omega\},
\end{aligned}
\end{equation}
where $\emptyset$ represents an empty set.

If $A \in 2^{\Omega}$, $A$ is called a proposition.
\end{myDef}

\begin{myDef}(Mass function)

A mass function $m$ in the frame of discernment $\Omega$ can be described as a mapping from $2^{\Omega}$ to [0, 1], defined as
\begin{equation}\label{eq_Massfunction1}
 m: \quad 2^{\Omega} \rightarrow [0, 1],
\end{equation}
satisfying the following conditions,
\begin{equation}
\label{eq_Massfunction2}
\begin{aligned}
 m(\emptyset) &= 0, \text{ and }
 \sum\limits_{A \in 2^{\Omega}} m(A) &= 1.
 \end{aligned}
\end{equation}
\end{myDef}

In the DS evidence theory, $m$ can also be called a basic belief assignment (BBA).
If $m(A)$ is greater than zero, where $A \in 2^{\Omega}$, $A$ is called a focal element.

\begin{myDef}(Belief function)

Let $A$ be a proposition in the frame of discernment $\Omega$.
The belief function of proposition $A$, denoted as $Bel(A)$ is defined by
\begin{equation}\label{eq_belieffunction}
\begin{aligned}
Bel(A) = \sum\limits_{B \subseteq A} m(B).
\end{aligned}
\end{equation}
\end{myDef}

\begin{myDef}(Plausibility function)

Let $A$ be a proposition in the frame of discernment $\Omega$.
The plausibility function of proposition $A$, denoted as $Pl(A)$ is defined by
\begin{equation}\label{eq_plausibilityfunction}
\begin{aligned}
Pl(A) = \sum\limits_{B \cap A \neq \emptyset} m(B).
\end{aligned}
\end{equation}
\end{myDef}

The belief function $Bel(A)$ and plausibility function $Pl(A)$ represent the lower and upper bound functions of the proposition $A$, respectively.

\section{The complex basic belief assignment~\cite{Xiao2018Generalization}}\label{CBBA}

A generalization of Dempster--Shafer evidence (GDSE) theory is presented recently, in which a new concept of complex belief function is defined based on the complex numbers~\cite{Xiao2018Generalization}.

Let $\Omega$ be a set of mutually exclusive and collective non-empty events, defined by
\begin{equation}\label{eq_Frameofdiscernment1}
 \Omega = \{e_{1}, e_{2}, \ldots, e_{i}, \ldots, e_{n}\},
\end{equation}
where $\Omega$ represents a frame of discernment.

The power set of $\Omega$ is denoted by $2^{\Omega}$, in which
\begin{equation}\label{eq_Frameofdiscernment2}
\begin{aligned}
 2^{\Omega} = \{\emptyset, \{e_{1}\}, \{e_{2}\}, \ldots, \{e_{n}\}, \{e_{1}, e_{2}\}, \ldots, \{e_{1}, \\
 e_{2}, \ldots, e_{i}\}, \ldots, \Omega\},
\end{aligned}
\end{equation}
and $\emptyset$ is an empty set.

\begin{myDef}(Complex mass function)\label{def_Complexmassfunction}

A complex mass function $\mathds{M}$ in the frame of discernment $\Omega$ is modeled as a complex number, which is represented as a mapping from $2^{\Omega}$ to $\mathbb{C}$, defined by
\begin{equation}\label{eq_GMassfunction1}
 \mathds{M}: \quad 2^{\Omega} \rightarrow \mathbb{C},
\end{equation}
satisfying the following conditions,
\begin{equation}\label{eq_Massfunction2}
\begin{aligned}
 \mathds{M}(\emptyset) &= 0, \\
 \mathds{M}(A) &= \mathbf{m}(A) e^{i \theta(A)}, \quad A \in 2^{\Omega} \\
 \sum\limits_{A \in 2^{\Omega}} \mathds{M}(A) &= 1,
\end{aligned}
\end{equation}
\end{myDef}
where $i = \sqrt{-1}$; $\mathbf{m}(A) \in [0, 1]$ representing the magnitude of the complex mass function $\mathds{M}(A)$;
$\theta(A) \in [-\pi, \pi]$ denoting a phase term.

In Eq.~(\ref{eq_Massfunction2}), $\mathds{M}(A)$ can also be expressed in the ``rectangular'' form or ``Cartesian'' form, denoted by
\begin{equation}\label{eq_Massfunction3}
\mathds{M}(A) = x + yi, \quad A \in 2^{\Omega}
\end{equation}
with
\begin{equation}\label{eq_Massfunction4}
\sqrt{x^2+y^2} \in [0, 1].
\end{equation}

By using the Euler's relation, the magnitude and phase of the complex mass function $\mathds{M}(A)$ can be expressed as
\begin{equation}\label{eq_magnitudeandphase}
\mathbf{m}(A)=\sqrt{x^2+y^2}, \text{ and } \theta(A) = \arctan(\frac{y}{x}),
\end{equation}
where $x = \mathbf{m}(A) \cos(\theta(A))$ and $y = \mathbf{m}(A) \sin(\theta(A))$.

The square of the absolute value for $\mathds{M}(A)$ is defined by
\begin{equation}\label{eq_squaremagnitude}
|\mathds{M}(A)|^2=\mathds{M}(A)\mathds{\overline{M}}(A)=x^2+y^2,
\end{equation}
where $\mathds{\overline{M}}(A)$ is the complex conjugate of $\mathds{M}(A)$, such that $\mathds{\overline{M}}(A)=x - yi$.

These relationships can be then obtained as
\begin{equation}\label{eq_relationship}
\mathbf{m}(A)=|\mathds{M}(A)|, \text{ and } \theta(A) = \angle \mathds{M}(A),
\end{equation}
where if $\mathds{M}(A)$ is a real number (i.e., $y = 0$), then $\mathbf{m}(A) = |x|$.

If $|\mathds{M}(A)|$ ($A \in 2^{\Omega}$) is greater than zero, $A$ is called a focal element of the complex mass function.
The value of $|\mathds{M}(A)|$ represents how strongly the evidence supports $A$.

The complex mass function $\mathds{M}$ modeled as a complex number in the generalized Dempster--Shafer (GDS) evidence theory can also be called a complex basic belief assignment (CBBA).
When $\mathds{M}(A)$ degrades into a real number, a CBBA will degrades into a BBA.

\begin{myDef}(Complex belief function)

Let $\Omega$ be a frame of discernment, and $A \in 2^\Omega$.
The complex belief function of $A$, denoted as $Bel_c(A)$ is defined by
\begin{equation}\label{eq_Gbelieffunction}
\begin{aligned}
Bel_c(A) = \sum\limits_{B \subseteq A} \mathds{M}(B).
\end{aligned}
\end{equation}
\end{myDef}

\begin{myDef}(Complex plausibility function)

Let $\Omega$ be a frame of discernment, and $A \in 2^\Omega$.
The complex plausibility function of $A$, denoted as $Pl_c(A)$ is defined by
\begin{equation}\label{eq_Gplausibilityfunction}
\begin{aligned}
Pl_c(A) = \sum\limits_{B \cap A \neq \emptyset} \mathds{M}(B).
\end{aligned}
\end{equation}
\end{myDef}

\section{A new distance measure between complex basic belief assignments}\label{Proposed method_EDM}
In this section, a new evidential distance measure for complex basic belief assignments is proposed.

\begin{myDef}(Evidential distance measure between CBBAs).
\label{def_EDMdistancemeasure}

Let $\mathds{M}_1$ and $\mathds{M}_2$ be two CBBAs on the frame of discernment $\Omega$, where $A$ and $B$ are the hypotheses of CBBAs $\mathds{M}_1$ and $\mathds{M}_2$, respectively.
The evidential distance measure between the CBBAs $\mathds{M}_1$ and $\mathds{M}_2$, denoted as $d_{CBBA}(\mathds{M}_1,\mathds{M}_2)$ is defined by
\begin{equation}\label{eq_distancemeasure1}
d_{CBBA}(\mathds{M}_1,\mathds{M}_2) =
\sqrt{\frac{|(\overrightarrow{\mathds{M}}_1-\overrightarrow{\mathds{M}}_2)^{T}
\underline{\underline{\mathds{D}}}
(\overrightarrow{\mathds{M}}_1-\overrightarrow{\mathds{M}}_2)|}{\sum\limits_{A \subseteq \Omega} |\mathds{M}_1(A)| + \sum\limits_{B \subseteq \Omega} |\mathds{M}_2(B)|}},
\end{equation}
where $\overrightarrow{\mathds{M}}$ is the vector of CBBA $\mathds{M}$;
$(\overrightarrow{\mathds{M}}_1-\overrightarrow{\mathds{M}}_2)^{T}$ is the transposition of $(\overrightarrow{\mathds{M}}_1-\overrightarrow{\mathds{M}}_2)$;
$|\cdot|$ denotes the absolute function;
$\underline{\underline{\mathds{D}}}$ represents a $2^n \times 2^n$ matrix which has the following elements
\begin{equation}
\underline{\underline{\mathds{D}}}(A,B) =
\frac{|A \cap B|}{|A \cup B|}.
\end{equation}

In Eq.~(\ref{eq_distancemeasure1}), $\sum\limits_{A \subseteq \Omega} |\mathds{M}_1(A)| + \sum\limits_{B \subseteq \Omega} |\mathds{M}_2(B)|$ is required to normalize $d_{CBBA}$.

For Eq.~(\ref{eq_distancemeasure1}), it can be expressed by another form,
\begin{equation}\label{eq_distancemeasure2}
d_{CBBA}(\mathds{M}_1,\mathds{M}_2) =
\sqrt{\frac{\|\overrightarrow{\mathds{M}}_1\|^2+\|\overrightarrow{\mathds{M}}_2\|^2-
2 |\langle\overrightarrow{\mathds{M}}_1,\overrightarrow{\mathds{M}}_2\rangle|}{\sum\limits_{A_i \in 2^\Omega} |\mathds{M}_1(A_i)| + \sum\limits_{A_j \in 2^\Omega} |\mathds{M}_2(A_j)|}},
\end{equation}
where $|\langle\overrightarrow{\mathds{M}}_1,\overrightarrow{\mathds{M}}_2\rangle|$ represents the scalar product, which is defined as
\begin{equation}\label{eq_distancemeasure21}
|\langle\overrightarrow{\mathds{M}}_1,\overrightarrow{\mathds{M}}_2\rangle| =
\left|\sum_{i=1}^{2^n} \sum_{j=1}^{2^n}
\mathds{M}_1(A_i) \mathds{\overline{M}}_2(A_j)
\frac{|A_i \cap A_j|}{|A_i \cup A_j|}\right|,
\end{equation}
$\mathds{\overline{M}}_2(A_j)$ is the complex conjugate of $\mathds{M}_2(A_j)$,
and $\|\overrightarrow{\mathds{M}}\|^2$ is the square norm of $\overrightarrow{\mathds{M}}$, defined by
\begin{equation}\label{eq_distancemeasure22}
\begin{split}
\|\overrightarrow{\mathds{M}}\|^2 &= |\langle\overrightarrow{\mathds{M}},\overrightarrow{\mathds{M}}\rangle| \\
&=
\left|\sum_{i=1}^{2^n} \sum_{j=1}^{2^n}
\mathds{M}(A_i) \mathds{\overline{M}}(A_j)
\frac{|A_i \cap A_j|}{|A_i \cup A_j|}\right|.
\end{split}
\end{equation}

\end{myDef}

It is obvious that when the CBBAs are degraded from complex numbers to real numbers, i.e., BBAs, the proposed distance measure degrades into the Jousselme et al.'s distance measure~\cite{jousselme2001new}.

The properties of the proposed distance measure can be summarized as 

\begin{Property}
Let $\mathds{M}_1$, $\mathds{M}_2$ and $\mathds{M}_3$ be arbitrary three CBBAs, then

P2.1 Non-negativity: $d_{CBBA}(\mathds{M}_1,\mathds{M}_2) \geq 0$.

P2.2 Non-degeneracy: $d_{CBBA}(\mathds{M}_1,\mathds{M}_2) = 0$ if and only if $\mathds{M}_1 = \mathds{M}_2$.

P2.3 Symmetry: $d_{CBBA}(\mathds{M}_1,\mathds{M}_2)$ = $d_{CBBA}(\mathds{M}_2,\mathds{M}_1)$.

P2.4 Triangle inequality: $d_{CBBA}(\mathds{M}_1,\mathds{M}_3) \leq d_{CBBA}(\mathds{M}_1,\mathds{M}_2) + d_{CBBA}(\mathds{M}_2,\mathds{M}_3)$.

P2.5 Boundedness: $0 \leq d_{CBBA}(\mathds{M}_1,\mathds{M}_2) \leq 1$.
\end{Property}

\section{Numerical examples}\label{Numericalexamples}

\begin{figure*}[htpb]
\centering
\subfigure[The comparison of different distance measures.]{
\begin{minipage}[c]{0.4\textwidth}
\centering
\includegraphics[width=7cm]{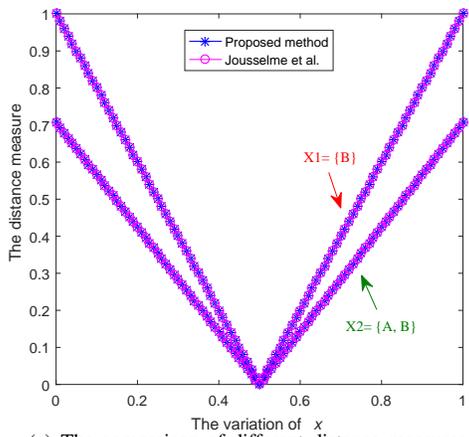}
\label{generalization11}
\end{minipage}}
\hspace{0.02\textwidth}
\subfigure[The EDM distance measures under the variation of $x$.]{
\begin{minipage}[c]{0.4\textwidth}
\centering
\includegraphics[width=7cm]{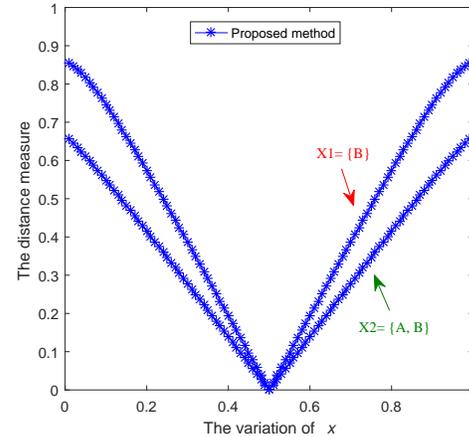}
\label{generalization12}
\end{minipage}}
\caption{The distance measures in Example~\ref{exa_generalization}.}
\label{generalization1}
\end{figure*}

\begin{exmp}\label{exa_generalization}
Assume there exist two CBBAs $\mathds{M}_1$ and $\mathds{M}_2$ in the frame of discernment $\Omega$:
\end{exmp}
\begin{equation*}
\begin{aligned}
\mathds{M}_1: & \quad
\mathds{M}_1(\{A\})=x+yi,
\mathds{M}_1(X_\theta)=1-x-yi; \\
\mathds{M}_2: & \quad
\mathds{M}_2(\{A\})=1-x+yi,
\mathds{M}_2(X_\theta)=x-yi;
\end{aligned}
\end{equation*}
where $\theta \in \{1,2\}$.
When $\theta=1$ and $\theta=2$, $X_1 = \{B\}$ and $X_2 = \{A,B\}$, respectively.
The belief values of $\mathds{M}_1$ and $\mathds{M}_2$ change as the variation of parameters $x$ and $y$.

When we set $y=0$, the CBBAs $\mathds{M}_1$ and $\mathds{M}_2$ degrade into real numbers.
The corresponding variation of the EDM distance measure between $\mathds{M}_1$ and $\mathds{M}_2$ is depicted in Fig.~\ref{generalization11} as $x$ varies within $[0, 1]$.
Meanwhile, the Jousselme et al.'s distance measure is also shown in Fig.~\ref{generalization11}.

From Fig.~\ref{generalization11}, we can notice that regardless of the singleton or multi-sets of $\mathds{M}_1$ and $\mathds{M}_2$, the EDM distance measure is exactly the same with the Jousselme et al.'s distance measure as $x$ changes within $[0, 1]$.
This result verifies that when $\mathds{M}_1$ and $\mathds{M}_2$ degrade into real numbers from complex numbers, the EDM distance measure degrades into the Jousselme et al.'s distance measure.

When we set $x$ within $[0.01, 0.99]$ and $y=0.1$, the CBBAs $\mathds{M}_1$ and $\mathds{M}_2$ are complex numbers.
The corresponding variations of EDM distance measures in terms of the singleton and multiple sets of $\mathds{M}_1$ and $\mathds{M}_2$ are depicted in Fig.~\ref{generalization12}, respectively, as $x$ changes within $[0.01, 0.99]$.

It can be noticed that when $x=0.5$ and $y=0.1$, the EDM distance measures are zero no matter $\mathds{M}_1$ and $\mathds{M}_2$ have singleton or multiple sets.
Whereas, for another cases that $y=0.1$ and $x$ within $[0.01, 0.5)\vee(0.5, 0.99]$, even $\mathds{M}_1$ and $\mathds{M}_2$ have the same belief values, the EDM distance measure between $\mathds{M}_1$ and $\mathds{M}_2$ under the case of $\theta=2$ with $X_\theta = \{A,B\}$ is smaller than that of the case that $\theta=1$ with $X_\theta = \{B\}$.
This result is reasonable and intuitive.
The reason is that under the case of $\theta=2$, there is an intersection $\{A\}$ between the subsets of $\mathds{M}_1$ and $\mathds{M}_2$, however, under the case of $\theta=1$, there is no intersection between the subsets of $\mathds{M}_1$ and $\mathds{M}_2$.

Moreover, from the results shown in Fig.~\ref{generalization11} and Fig.~\ref{generalization12}, it is obvious that the symmetry property of EDM distance measure is verified as well as the non-negativeness property.

\section{Conclusions}\label{Conclusion}
In this paper, a new distance measure is proposed for complex basic belief assignments, called as EDM distance measure in a more general framework of complex plane space.
In addition, the properties of the proposed EDM distance measure are defined and analyzed.
It is proofed that the EDM distance is a strict distance metric, as it satisfies the axioms of a distance.
Then, numerical examples illustrate the effectiveness of the EDM distance measure.

The main contribution of this study is that the EDM distance measure is a generalization of the Jousselme et al.'s distance measure.
In particular, when the complex basic belief assignments become basic belief assignments, the EDM distance measure degrades into Jousselme et al.'s distance measure in evidence theory.
In summary, this study is the first work to consider the distance measure between evidences in the framework of complex numbers.
It provides a promising way to measure the difference or dissimilarity in the process of solving the decision-making problems.

\section*{Acknowledgment}
This research is supported by the Fundamental Research Funds for the Central Universities (No. XDJK2019C085) and Chongqing Overseas Scholars Innovation Program (No. cx2018077).

\ifCLASSOPTIONcaptionsoff
  \newpage
\fi

\bibliographystyle{IEEEtran}
\end{document}